%% file: videoLM_iclr2015.tex
\documentclass{article} 
\usepackage{iclr2015,times}
\usepackage{hyperref}
\usepackage{url}
\usepackage{graphicx}
\usepackage[cmex10]{amsmath}
\usepackage{array}
\usepackage{mdwmath}
\usepackage{mdwtab}
\usepackage{eqparbox}
\usepackage{stfloats}
\usepackage{times}
\usepackage{epsfig}
\usepackage{amssymb}

\title{Video (Language) Modeling:  A Baseline \\For Generative Models of Natural Videos}

\author{
Marc'Aurelio Ranzato$^1$,  Arthur Szlam$^1$, Joan Bruna$^{1,2}$, Michael Mathieu$^{1,2}$, Ronan Collobert$^1$, Sumit Chopra$^1$ \\
$^1$Facebook AI Research  \hspace{2cm} $^2$Courant Institute of Mathematical Sciences\\
Menlo Park CA \& New York NY USA \\
\texttt{\{ranzato,aszlam,locronan,spchopra\}@fb.com} \hspace{.5cm} \texttt{\{bruna, mathieu\}@cims.nyu.edu} \\
}

\newcommand{\bh}{\mathbf{h}} 
\newcommand{\bb}{\mathbf{b}}
 
\newcommand{\bone}{\mathbf{1}}

\iclrconference 

\begin{document}

\maketitle

\begin{abstract}
We propose a strong baseline model for unsupervised feature learning 
using video data.   By learning to predict missing frames or extrapolate future frames from an
input video sequence, the model discovers both spatial and temporal correlations
which are useful to represent complex deformations and motion patterns.
The models we propose are largely borrowed from the language modeling literature, and adapted
to the vision domain by quantizing the space of image patches into a large dictionary.

We demonstrate the approach on both a filling and a generation task. For the first time,
we show that, after training on natural videos, 
 such a model can predict non-trivial motions over short video sequences.
\end{abstract}

\section{Introduction} \label{introduction}
The human visual system has the ability to learn new visual concepts from only a handful of instances.  
It can  generalize to new views, new lighting conditions, and it is robust to large within-class variations. 
Arguably, a key property of a good artificial visual system is to also be able to discover patterns on-the-fly and 
to generalize from very few labeled instances. Therefore, it seems natural to investigate computational models that 
can leverage unlabeled data to discover regularities and structure from the visual world without using any annotation. 

Despite great advances in object recognition, detection, and parsing over the past few
 years~\citep{Krizhevsky:NIPS12,hariharan-eccv2014,simonyan-nips2014,girshick2014rcnn}, 
   none of the widely used methods for these tasks relies on unlabeled data. Instead, they require very large and carefully annotated datasets.
 There is a vast body of literature on unsupervised learning for vision, for example ~\citep{Hinton-DeepAutoencoder,vincent-icml08,psd},
 but these methods have not found success in practical applications yet. 

The biggest hurdle to overcome when learning without supervision is the design of an objective function that encourages the system
 to discover meaningful regularities. One popular objective is squared Euclidean distance between the input and its reconstruction
 from some extracted features. Unfortunately, the squared Euclidean distance in pixel space is not a good metric, 
since it is not stable to small image deformations, and responds to uncertainty with linear blurring.
Another popular objective is log-likelihood, reducing unsupervised learning to a density estimation problem. 
However, estimating densities in very high dimensional spaces can be difficult, particularly 
the distribution of natural images which is highly  concentrated and multimodal. 

Although some recent works have shown some progress on smaller resolution
 images~\citep{goodfellow-nips2014, zoran-2012, ranzato-pami2013, theis-2011} using generative models,
 it is unclear how these models can scale to handle full resolution inputs, due to the curse of dimensionality.
 Other models attempt to overcome this problem by using priors over the features, such as
 sparsity~\citep{Olshausen-Field,psd,lee-nips-07}. Although these constraints make the learning problem better posed, 
they are still too simplistic to capture complex interactions between features.   
While several authors have reported realistic generation of small image patches, fewer works have operated 
at the scale of high-resolution images~\citep{Theis2012a, ranzato-pami2013} and success has been more limited.

On the other hand, many have argued that learning without (human) supervision can become much easier once we consider
 not just a collection of independently drawn natural images, but a dataset of natural videos~\citep{ostrovsky-2009}.  
Then, spatial-temporal correlations can provide powerful information about how objects deform, about occlusion, object boundaries,
 depth, and so on. By just looking at a patch at the same spatial location across consecutive time steps, the system can infer
 what the relevant invariances and local deformations are. Even more so when studying generative models of natural images, 
modeling becomes easier when conditioning on the previous frame as opposed to unconditional generation, 
yet this task is non-trivial and useful as the model has to understand how to propagate motion and cope with occlusion.

Research on models that learn unsupervised from videos is still in its infancy. In their seminal work,
 \citet{hateren98} and \citet{hurri03} have applied ICA techniques to small video cubes of patches. 
\citet{SFA} have proposed instead a method based on slowness of features through time, an idea later extended 
in a bilinear model by~\citet{gregor_lc10}. 
Bilinear models of transformations between nearby frames have also been investigated 
by~\citet{MemisevicHinton09,sutskever09,taylor-jmlr11,memisevic-nips2014} 
as well as~\citet{rao-2007} via Lie group theory. Related work by~\citet{cadieu-09} uses a hierarchical model with
 a predefined decomposition between amplitude (which varies slowly) and phase (encoding actual transformations). 
Perhaps with the only exception given by the layered model proposed by ~\citet{sprites01}, 
all the above mentioned models have been demonstrated on either small image patches or small synthetic
 datasets~\citep{sutskever09, sahani-nips12, memisevic-nips2014}. 
One major drawback of these models is that they are often not easy to extend  to large frames, they do not scale to 
datasets larger than a few thousand frames and they do not generalize to a large number of generic transformations.

In this work, we propose a stronger baseline for video modeling. Inspired by work in the language modeling 
community~\citep{nn-lm, rnn-lmm}, we propose a method that is very simple yet very effective, 
as it can be applied to full resolution videos at a modest computational cost. 
The only assumption that we make is local spatial and temporal stationarity of the input 
(in other words, we replicate the model and share parameters both across space and time), 
but no assumption is made on what features to use, nor on how to model the transformation between nearby frames. 
Despite our rather simplistic assumptions, we show that the model is actually able to capture non-trivial deformations. 
To the best of our knowledge, we demonstrate for the first time that a parametric model can generate realistic predictions 
of short video sequences after being trained on natural videos.

\input{model.tex}

\input{experiments.tex}

\input{conclusion.tex}

{\small
\bibliography{videoLM_iclr2015}
\bibliographystyle{iclr2015}
}
\newpage
\input{appendix.tex}

\end{document}

%% file: model.tex
\section{Model} \label{model}
Given a sequence of consecutive frames from a video, denoted by $(X_1, X_2, \dots, X_t)$,   we might want to train a system to predict the next frame in the sequence, $X_{t+1}$, where the subscript denotes time.   More generally, given some context of frames, we might want to try to predict some frames that have been left out of the context. 
This is a simple and well defined task which does not require labels, yet accurate predictions can only be produced by models that have learned motion primitives and understand the local deformations of objects.  
At test time, we can validate our models on both generation and filling tasks (see sec.~\ref{generation} and \ref{filling}).    

In order to design a system that tackles these tasks, we draw inspiration from classical methods from 
natural language processing, namely n-grams, neural net language models~\citep{nn-lm} 
and recurrent neural networks~\citep{rnn-lmm}. 
We will first review these methods and then explain how they can be extended to model video sequences (as opposed to sequences of words).

\subsection{Language Modeling} \label{LM}
In language modeling, we are given a sequence of {\em discrete} symbols (e.g., words) from a finite (but very large) dictionary. Let a symbol in the sequence be denoted by $X_k$ (symbol at position $k$ in the sequence); if $V$ is the size of the dictionary, $X_k$ is an integer in the range $[1, V]$.

In language modeling, we are interested in computing the probability of a sequence of words, $p(X_1, X_2, \dots, X_N) = p(X_N | X_{N-1}, \dots, X_{1}) p(X_{N-1} | X_{N-2}, \dots, X_{1}) \dots p(X_2 | X_1) p(X_1)$. Therefore, everything reduces to computing the conditional distribution: $p(X_k | X_{k-1}, \dots, X_{1})$. In the following, we will briefly review three methods to estimate these quantities. 

\subsubsection{n-gram} \label{ngram}
The {\em n-gram} is a table of normalized frequency counts under the Markovian assumption that
$p(X_t | X_{t-1}, \dots\, X_{1}) = p(X_t | X_{t-1}, \dots, X_{t - n + 1})$. In this work, these conditional probabilities are computed by the count ratio:
\begin{equation}
p(X_t | X_{t-1}, \dots, X_{t - n + 1}) =  \frac{count(X_{t - n + 1}, \dots, X_t) + 1}{count(X_{t - n + 1}, \dots, X_{t-1}) + V} \label{eq:ngram}
\end{equation}
where the constants in the numerator and denominator are designed to {\em smooth} the distribution and improve generalization on unfrequent n-grams (Laplace smoothing). In this work, we considered bigrams and trigrams (n=2 and n=3, respectively).
 
\subsubsection{Neural Net Language Model} \label{nnet}
The neural net language model (NN)~\citep{nn-lm} is a parametric and non-linear extension of n-grams. Let $\bone(X_k)$ be the 1-hot vector representation of the integer $X_k$, that is, a vector with all entries set to 0 except the $X_k$-th component which is set to 1. 
In this model, the words in the context (those upon which we condition) are first transformed into their 1-hot vector representation, 
they are then linearly embedded using matrix $W_x$, the embeddings are concatenated and 
finally fed to a standard multilayer neural network. This network is trained using a cross-entropy loss to predict the next word in the sequence (usual multi-class classification task with $V$ classes). In this model, the output probability is then given by:
\begin{equation}
  p(X_t | X_{t-1}, \dots, X_{t - n + 1}) =  SM(MLP[W_x \bone(X_{t-1}), \dots, W_x \bone(X_{t-n+1})])\label{eq:nnet}
\end{equation}
where $SM$ stands for softmax and MLP any multi-layer neural network (in our case, it is a one hidden layer neural network with ReLU units).
Note that the first layer of MLP acts like a look up table (given the input encoding), mapping each discrete symbol into a
 continuous embedding vector.

\subsubsection{Recurrent Neural Network} \label{rnn}
The recurrent neural network (rNN)~\citep{rnn-lmm}, works similarly to the model above except that: 
1) it takes only one input at the time, and 
2) the hidden layer of the MLP takes as input also a linear transformation of the hidden state at the previous time step. 
This enables the rNN to potentially leverage a variable size context without compromising computational efficiency. The equations that regulate the rNN are:
\begin{eqnarray}
\bh_{t - 1} = \sigma(W_h \bh_{t - 2} + W_x \bone(X_{t - 1})), \hspace{.4cm} 
p(X_t | \bh_{t - 1}) =  SM(W_o \bh_{t - 1} + \bb_o) \label{eq:rnn}
\end{eqnarray}
Training the parameters of the model, $\{W_x, W_h, W_o, \bb_o\}$, proceeds by minimization of the standard cross-entropy loss on the next symbol using back-propagation 
through time~\citep{autoencoder-nn} and gradient clipping~\citep{rnn-lmm}.

\subsection{Video (Language) Modeling}
The above mentioned methods work on a sequence of discrete input values; however, video frames are usually received as continuous vectors (to the extent that 8-bit numbers are continuous).
If we want to use these methods to process video sequences, we can follow two main strategies. 
We can either replace the cross-entropy loss with mean squared error (or some other regression loss), or we can discretize the frames. 

The former approach turns the classification problem into regression. As mentioned in sec.~\ref{introduction}, 
this is hard because it is very easy for the model to produce relatively low reconstruction errors by merely blurring the last frame. In our experiments, we found that this approach was harder to optimize and yielded results only marginally better than simply predicting the last frame (relative MSE improvement of 20\% only).

The other option is to discretize the input, for instance  by using k-means. 
 Instead of operating in the pixel space
where we do not know a good metric for comparing image patches, we operate in a very sparse feature space, where each patch is 
coded by a $k$-means atom.  This sparsity enforces strong
constraints on what is a feasible reconstruction, as the $k$-means atoms ``parameterize'' the space of outputs.    The prediction problem
 is then simpler because the {\it video model} does not have to parameterize the output space; it only has to decide where in the output space the next prediction should go.
On the other hand, with even a small set of centroids, there are a huge number of possible combinations of centroids that could reasonably occur in an image or video sequence, and so the prediction problem is still non-trivial. 
There is clearly a trade-off between quantization error and temporal prediction error.
The larger the quantization error (the fewer the number of centroids), the easier it will be to predict the codes for the next frame, and 
vice versa. 
In this work, we quantize small gray-scale 8$\times$8 patches using 10,000 centroids constructed via $k$-means, and represent an image as a 2$d$ array indexing the centroids.

\begin{figure}[!t]
\begin{center}
\vspace{-.8cm}
\includegraphics[width=0.2\linewidth]{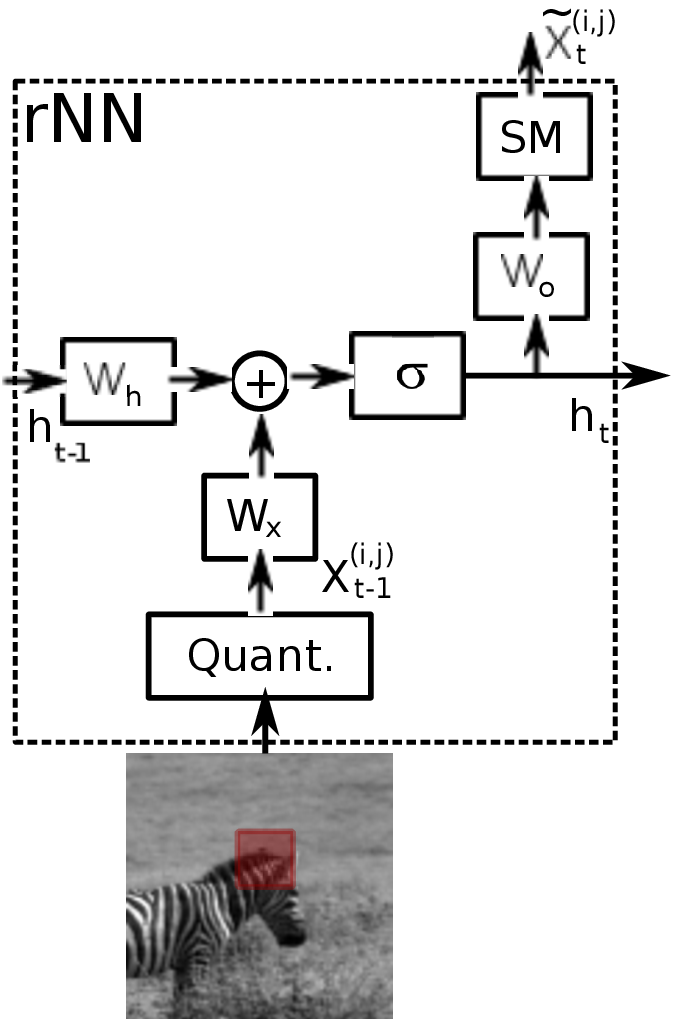}
\hspace{.2cm}
\includegraphics[width=0.2\linewidth]{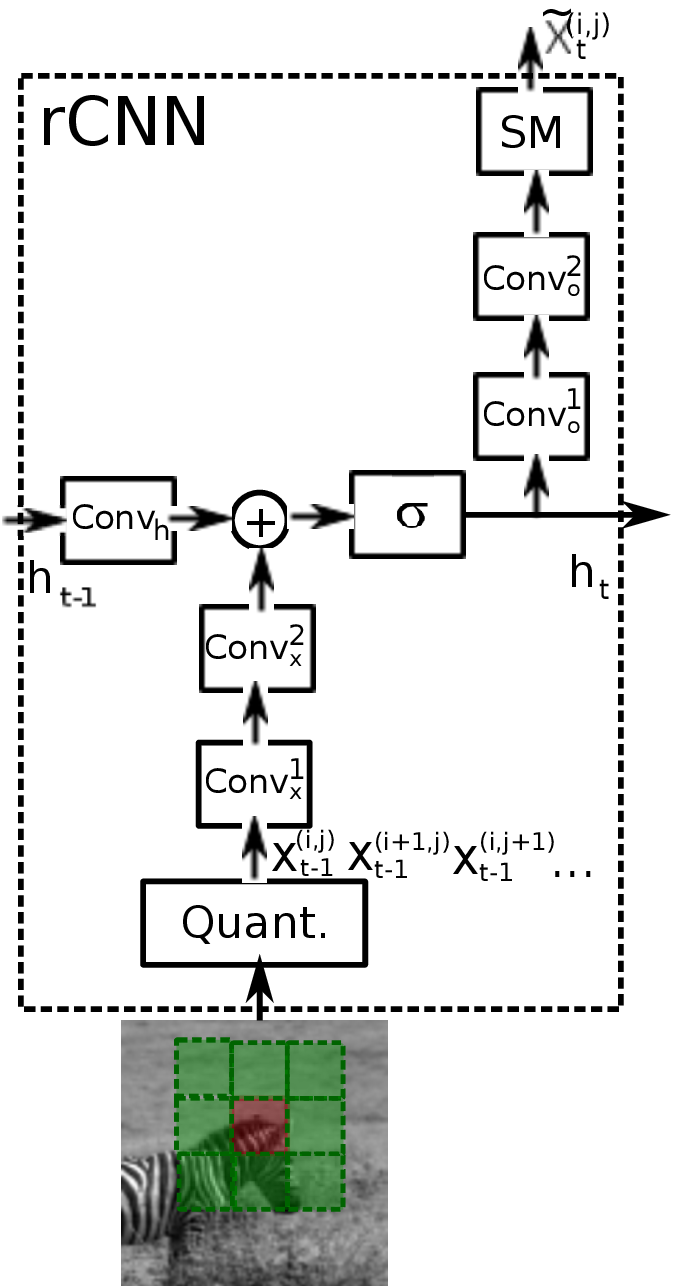}
\hspace{.2cm}
\includegraphics[width=0.2\linewidth]{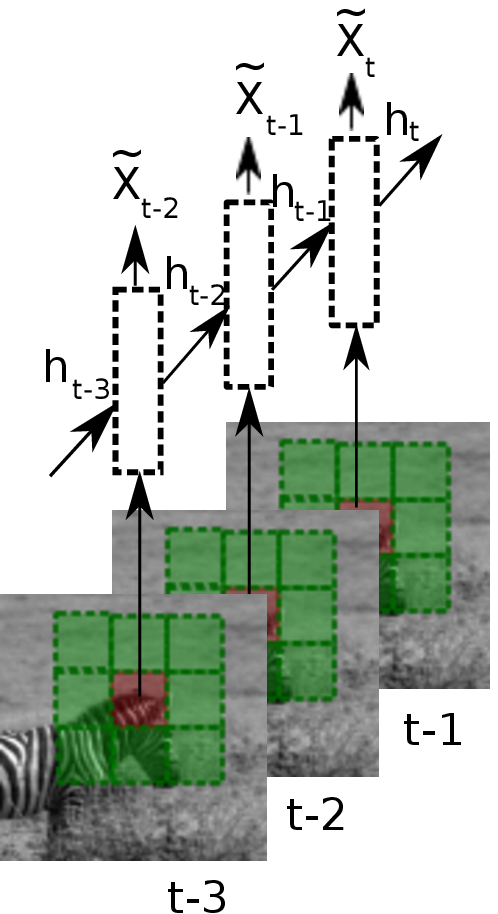}
\end{center}
\vspace{-.2cm}
\caption{Left: outline of an rNN building block applied to video patch modeling. Center: outline of rCNN building block applied to 
frame-level video modeling. The rCNN takes as input a patch of quantized patches and uses both spatial and temporal context 
to predict the central patch at the next time step. At test time, it can be unrolled spatially over any frame size.
Right: example of how such blocks are replicated over time to model a video sequence 
(sharing parameters over time).}
\label{fig:outline_rnn}
\end{figure}

To summarize, we apply the language modeling methods described above by quantizing video frames using $k$-means on non-overlapping image patches. When modeling video cubes of patches (i.e., patches at the same location but across consecutive time steps, 
see fig.~\ref{fig:outline_rnn} left), the approach is rather straightforward and will be empirically evaluated in sec.~\ref{zebra}.

\vspace{-.2cm}

\subsubsection{Recurrent Convolutional Neural Network}
The last model we propose is a simple extension of the rNN to better handle spatial correlations between nearby image patches. 
In the rNN, nearby patches are treated independently while there are almost always very strong spatial correlations between nearby patches.
In the recurrent convolutional neural network (rCNN) we therefore feed the system with not only a single patch, but also with the nearby
patches. The model will not only leverage temporal dependencies but also spatial correlations to more accurately predict the central patch
at the next time step (see fig.~\ref{fig:outline_rnn} center).

The rCNN that we will use in our experiments, takes as input a patch of size 9$\times$9. Each integer in the grid corresponds to a 
 quantized patch of size 8$\times$8 pixels. In this work, these patches do not overlap although everything we describe would apply to
 overlapping patches as well. This input patch of integers is first embedded into a continuous feature space as in a standard rNN (matrix
$W_x$ in eq.~\ref{eq:rnn}),
 and then passed through two convolutional layers. In fig.~\ref{fig:outline_rnn} center, ``$Conv_x^1$'' actually represents: embedding
followed by convolution and logistic non-linearity. All convolutional layers use 128 filters of size 3$\times$3. Because of border effects,
the recurrent code has 128 feature maps of size 5$\times$5. To avoid border effects in the recurrent code (which could propagate in time
with deleterious effects), the transformation between the recurrent code at one time step and the next one is performed
by using 1$\times$1 convolutional filters (effectively, by using a fully connected layer which is shared across all spatial locations).
Finally, the recurrent code is {\em decoded} through other two convolutional layers with 128 filters of size 3$\times$3.
These produce a vector of size 128 and spatial size 1$\times$1 which is fed to a fully connected layer with $V$ outputs
 (in our case, 10,000).

At generation (test) time, we unroll the rCNN on a larger frame (since all layers are convolutional\footnote{
Fully connected layers can be interpreted as limit case of convolutional layers with kernels of size 1$\times$1.}).
The use of several convolutional layers in the decoder is a good guarantee that nearby predictions are going to be spatially consistent
because most of the computation is shared across them. Even though the recurrent code can fluctuate rapidly in response to a rapidly
varying input dynamics, the prediction is going to favor spatially coherent regions.

%% file: experiments.tex
\section{Experiments} \label{experiments}
In this section, we empirically validate the language modeling techniques
discussed in sec.~\ref{model}. Quantitatively, we evaluate models in terms of their 
ability to predict patches one frame ahead of time. 
We also show examples of generation and filling of short video
sequences that capture non trivial deformations.

In our experiments, we used the following training protocol.
First, we do not pre-process the data in any way except for gray-scale conversion and division
by the standard deviation. Second, we use 10,000 centroids for $k$-means quantization.
Third, we cross-validate all our hyper-parameters on the validation set and report accuracy
on the test set using the model that gave the best accuracy on the validation set.
For the van Hateren's dataset, we used 3 videos for validation and 3 videos for testing (out
of the 56 available). For the UCF 101 dataset, we used the standard split~\citep{ucf101}. Results
on the van Hateren's dataset are reported in the Supplementary Material for lack of space.

\begin{table}[!t]
\vspace{-.8cm}
\caption{Entropy in bits/patch and (perplexity) for predicting the
 next 8$\times$8 quantized patch (10,000 atoms in the dictionary). Left: van Hateren dataset. 
Right: UCF 101 dataset.}
\label{tab:logprob}
    \begin{tabular}{l||l|l|l}
      \multicolumn{1}{c}{\bf Model}  &\multicolumn{1}{c}{\bf Training} &
      \multicolumn{1}{c}{\bf Validation} &\multicolumn{1}{c}{\bf Test} \\
      \hline
      \hline
      bi-gram & 8.3 (314) & 9.9 (884) & 9.3 (647) \\
      \hline
      NN & 5.9 (59) & 7.6 (192) & 7.2 (146) \\
      \hline
      rNN & 5.9 (59) & 7.7 (211) & 7.3 (156) \\
    \end{tabular}
\quad
  \begin{tabular}{l||l|l|l}
      \multicolumn{1}{c}{\bf Model}  &\multicolumn{1}{c}{\bf Training} 
      & \multicolumn{1}{c}{\bf Validation} &\multicolumn{1}{c}{\bf Test} \\
      \hline
      \hline
      bi-gram & 4.8 (27) & 4.8 (27) & 4.9 (29) \\
      \hline
      NN & 4.4 (21) & 4.4 (21) & 4.5 (22) \\
      \hline
      rNN & 4.0 (16) & 4.3 (20) & 4.3 (20) \\
      \hline
      rCNN & 3.7 (13) & 3.8 (14) & 3.9 (15)
  \end{tabular}
\end{table}

\subsection{UCF-101 Dataset} \label{ucf101}
The UCF-101 dataset~\citep{ucf101} is a standard benchmark dataset for human action recognition.
It has 13320 videos of variable length belonging to 101 human action categories, and each frame has size 160$\times$320 pixels.
This dataset is interesting also for unsupervised learning because: 
a) it is much larger than the van Hateren dataset, and
b) it is much more realistic since the motion and the spatial scale of objects have not
been normalized. This dataset is by no means ideal for learning motion patterns either, 
since many videos exhibit jpeg artifacts and duplicate frames due to compression, which further complicate learning.

Tab.~\ref{tab:logprob} (right) compares several models. In this case, bigger models worked
generally better. In particular, the rCNN yields the best results, showing that not only the
temporal but also the spatial context are important for predicting a patch at the next time step.
The best rCNN was trained by using: 8 back-propagation through time steps, mini-batch size 128, 
learning rate set to 0.005 with momentum set to 0.9, and it had 128 feature maps at the output
 of each convolutional layer.

In order to understand how much quantization hurts generation and to make our results comparable to methods that 
directly regress pixel values (for which entropy and perplexity do not apply), 
we also report the average root mean square error (RMSE) per pixel value 
(using pixel values scaled in the range from 0 to 255). 
On the test set, rCNN achieves an average RMSE of 15.1 while a perfect model of the temporal dynamics 
(i.e., accounting only for the quantization error) gets an RMSE of 8.9.
Although RMSE is not a good metric to measure similarity between images, 
we can conclude that quantization error accounts for about half of the total error and
that temporal dynamics are captured fairly accurately (in average). In order to compute RMSE, we reconstructed
each patch using the most likely dictionary element and we averaged predictions for all 8$\times$8 spatial displacements, so that
each pixel is the average of 64 values. Averaging over spatial displacements was important to remove blocking artifacts due to the non-overlapping
nature of the quantization used.

\subsection{Analyzing the Model} \label{visualizations}
\begin{figure}[!t]
\begin{center}
\vspace{-.8cm}
\includegraphics[width=0.8\linewidth]{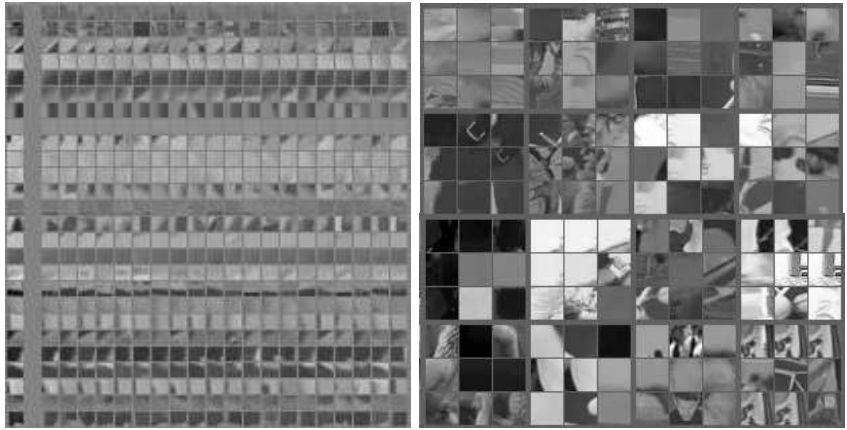}
\end{center}
\vspace{-.3cm}
\caption{Left: Example of embeddings learned by rCNN. The first column shows the $k$-means centroid
corresponding to a certain atom in the dictionary. The other columns show the centroids
 whose corresponding embeddings are the nearest neighbors to the one in the first
column. Each row corresponds to a randomly picked atom. Right: Each 3$\times$3 grid of patches
shows those input patches (from the validation set) that make a certain unit (output of 
the first convolutional layer) fire the most. Only a random subset of these units and embeddings are
shown.}
\vspace{-.4cm}
\label{fig:visualizationUCF101}
\end{figure}
\begin{table}[!t]
\caption{Analysis of the static (left) and dynamic (right) part of rCNN.
Perplexities are computed on a subset of UCF 101 validation set.
See main text for details.}
\label{tab:sttic_dynamic}
    \begin{tabular}{l||l}
      \multicolumn{1}{c}{\bf Model}  &\multicolumn{1}{c}{\bf Perplexity} \\
      \hline
      \hline
      1 frame (copy of previous), natural layout & 2.1 \\
      \hline
      static video (long squence), natural layout & 1.3\\
      \hline
      1 frame (copy of previous), random layout & 6.6 \\
      \hline
      static video (long squence), random layout & 2.0
    \end{tabular}
\quad
  \begin{tabular}{l||l}
      \multicolumn{1}{c}{\bf Model}  &\multicolumn{1}{c}{\bf Perplexity}\\
      \hline
      \hline
      natural & 12.2 \\
      \hline
      reversed & 12.3 \\
      \hline
      random & 100,000+ \\
      \hline
      skip 1 frame & 30
  \end{tabular}
\end{table}
In this section, we investigate what the rCNN has learned after training on the UCF~101 dataset.
First, we analyze the parameters in the embedding matrix and first convolutional layer. 

There is one embedding per
$k$-means centroid and we look at the centroids corresponding to the nearest neighbor
 embeddings. Fig.~\ref{fig:visualizationUCF101} (left) shows that the rCNN, but similarly the 
rNN and NN, learns to cluster together similar centroids. This means that, despite the
quantization step, the model learns to become robust to small distortions. It does not matter
much which particular centroid we pick, the vector that we output is going to be nearby
in space for similar looking input patches. 

We also visualize (a random subset of) the first layer convolutional filters by looking
at the input examples (from the validation set) that make the output fire the most.
Fig.~\ref{fig:visualizationUCF101} (right) shows that these patterns exhibit similar structure
but at slightly different position, orientation and scale. 

Finally, we try to disentangle the static (only spatial) and the dynamic (only temporal) 
part of rCNN.
In the left part of tab.~\ref{tab:sttic_dynamic} we compute the model's score for static video
sequences (initializing on a given frame) of different length. rCNN assigns high likelihood
to non-moving sequences. However, if we randomly permute the order of the patches (and maintain
such order for the whole sequence), the likelihood is lower - demonstrating a preference for
natural videos. This experiment show that rCNN does take into account the spatial context and
that it does not learn a mere identity function. The right part of the table investigates
the temporal part of the model. rCNN is barely able to distinguish between video sequences
that are played in the natural versus reversed order, but it definitely prefers temporally ordered
video sequences.

\subsection{Generation} \label{generation}
After training on UCF-101, we used the rCNN to predict future frames after conditioning upon 12
real consecutive frames. Generation proceeds as follows: a) we unroll the rCNN on whatever
frame size we use at test time and run it forward on all
the frames we condition upon, b) we take the most likely 
predicted atom as next input, and iterate. In order to alleviate quantization errors, we
perform the same procedure on all 64 spatial offsets (combination of 8 horizontal and 8 vertical
shifts) and then average the predictions for each pixel.

\begin{figure}[!t]
\begin{center}
\vspace{-1cm}
\includegraphics[width=0.9\linewidth]{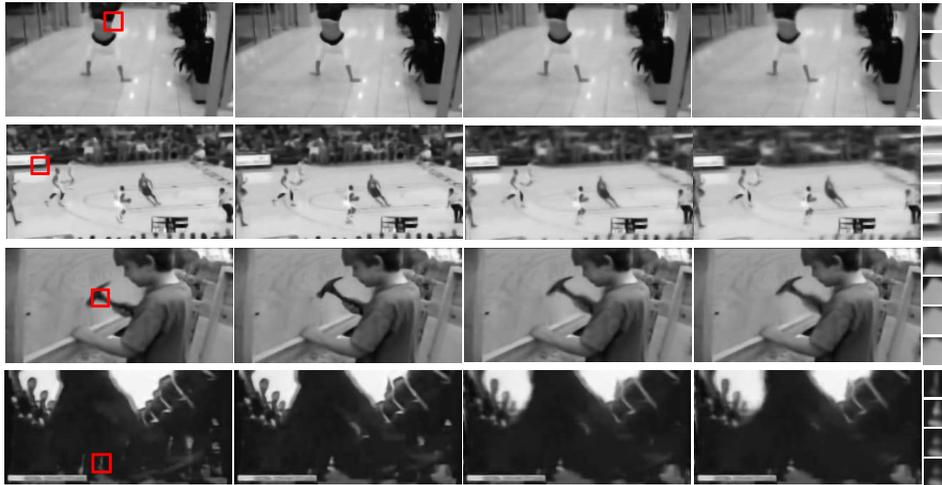}
\end{center}
\vspace{-.5cm}
\caption{Examples of generations after training on the UCF 101 dataset.
The first two frames (columns)
are used for initialization of the rCNN, the next two frames are
generations from rCNN. The last column shows a zoom of the 20$\times$20 pixel patch
marked in red (first frame on the top). Frames have size 160$\times$320 pixels.
More examples at: \tiny{\url{https://research.facebook.com/publications/video-language-modeling-a-baseline-for-generative-models-of-natural-videos/}}}
\label{fig:generation_demo}
\end{figure}

Fig.~\ref{fig:generation_demo} (right) shows that rCNN is fairly good at completing the motion,
even capturing fairly complex out of plane rotations and deformations. However, the predictions
tend to slow down motion quite rapidly, and eventually the model converges to a still image after a couple of frames in average.
Animations, longer generations and comparisons are available at {\tiny\url{https://research.facebook.com/publications/video-language-modeling-a-baseline-for-generative-models-of-natural-videos/}}.
Generally speaking, the model is good at predicting motion of fairly fast moving objects of
large size, but it has trouble completing videos with small or slowly moving objects.

In the url above, we also compare to the generation produced by a baseline method based on
optical flow. This estimates the flow on the first two frames, and applies it to the remaining
ones. This method exploits knowledge of low level image structure and local distorions. Although,
this baseline may produce good predictions of the next frame, 
it also degrades the quality of subsequent frames by introducing significant smearing artifacts.

\subsection{Filling} \label{filling}
We also evaluate a neural network language model on the task of filling in frames from a video, given boundary values.   
For simplicity, the boundary values include the top/bottom/left/right borders of the whole video cube (15 pixels wide), 
in addition to the frames used for temporal context (both past and future).   
We use a model which takes as input two 3$\times$3 patches of atoms at the same location from frames $j$ and $j+2$, 
and it is trained to predict the middle atom in the corresponding 3$\times$3 patch of the ($j+1$)-th frame.
   
\begin{figure}[!t]
\begin{center}
\vspace{-.2cm}
\includegraphics[width=0.6\linewidth]{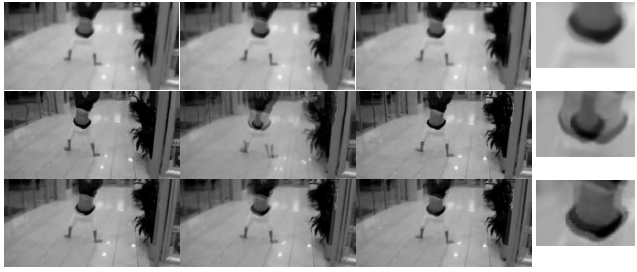}
\end{center}
\vspace{-.5cm}
\caption{Examples of 3 frames filled in by our algorithm, optical flow and linear interpolation (from top
to bottom). Time goes from left to right. On the left, we show a zoom of a patch.}
\label{fig:filling_demo}
\end{figure}

At test time, we use an iterative procedure to fill in larger gaps. At each iteration and spatio-temporal location $z$,
we take our current estimate of the spatio-temporal context $C$ of $z$,  and use it as input to our language model to
re-estimate the atom at $z$. 
In our experiments, we update each spatio-temporal location before updating the contexts and the iteration counter.
Finally, we reconstruct the quantized frames by averaging over the 64 shifts of the model as before.
Fig.~\ref{fig:filling_demo} shows examples of filling in $3$ consecutive frames from a UCF-101 video in the validation set.
The model compares favorably against both linear interpolation and an optical flow baseline.

Note that unsurprisingly, the problem of filling a single frame given its temporal neighbors is easier than extrapolating to the future. 
For example, the simple model described here achieves a validation entropy of $2.8$ bits/patch; compare to Table \ref{tab:logprob}.

%% file: conclusion.tex
\section{Discussion} \label{limitation}
Despite the promising results reported in the previous section, language modeling based techniques have also
several limitations, which may open avenues of future work.
\newline
{\bf Multi-Scale Prediction:} Multiscale, coarse-to-fine approaches are classic in motion estimation models~\citep{brox2004high}.
Similar approaches could be easily adopted for language modeling in the context of video, and in particular, for rCNN. 
For instance, one could use a standard multi-resolution pyramid, whereby a finer resolution level is fed with the residual errors
produced by a coarser resolution level. Moreover, since motion statistics are roughly scale invariant, 
the same rCNN could be used at all resolutions (except for the coarsest one, 
which operates on the local means rather than the local differences). Such scheme would allow to better model motion, regardless
of the speed and size of objects.
\newline
{\bf Multi-Step Prediction:} One of the limitations of the current model is that it cannot perform predictions further than a few frames 
ahead into the future. Without injecting sampling noise in the system, the model converges rapidly to a static image
because a) in average (over all the frames in the data) there is little if any motion and 
b) the peak of the distribution (recall that we propagate the max) does not capture its variance (i.e., how uncertain
the model is about the position of a certain edge). In particular, the fact that the distribution of natural videos has a strong
bias for non-moving inputs indicates that this task is intrinsically different from the usual language modeling one.  On the other hand,
injecting sampling noise independently at each spatial location hampers spatial coherence (see supplementary material). 
\newline
Although we do not have a full understanding of this issue, we conjecture that
one way to combat this phenomenon is to predict several steps ahead of time, feeding
the system with its own predictions. This will have several benefits: it will encourage the system to produce 
longer sequences and, at the same time, it will make the system robust against its own prediction errors. 
\newline
Another strategy is to modify the inference at generation time. Although running full Viterbi decoding is prohibitive 
due to the large number of spatio-temporal interaction terms, one could modify the greedy generation algorithm 
to take into account the joint spatio-temporal co-occurrences 
of image patches, for instance with n-grams over temporal and spatial slices.
\newline
{\bf Structured Prediction versus Regression:}
While we found it difficult to directly regress output frames in the $\l_2$ metric, quantization also introduces some drawbacks. 
Besides visual artifacts, it  makes the learning task harder 
than necessary, because all targets are assumed to be equally dissimilar, even though they are not. Moreover, quantization makes it hard
to learn end-to-end the whole system, to back-propagate easily through a multi-step prediction model, and to efficiently perform
joint inference of all patches in a given frame (given the combinatorial nature of the discrete problem).
\newline
{\bf Implicit VS Explicit Modeling of Transformations:}
The model we proposed does not have any explicit representation of transformations. It is therefore difficult to generate
perpetual motion, to extract motion features and to relate objects characterized by similar motion (or vice versa, to tell
whether the same object is moving in a different way). The ``what'' and ``where'' are entangled.
However, it seems straightforward to extend the proposed model to account for motion specific features. For instance, part
of the learned representation could be tied across frames to encode the ``what'', while the rest could be dedicated to represent
transformations, perhaps using bilinear models.

\section{Conclusion} \label{conclusion}
We have presented a baseline for unsupervised feature learning inspired by standard language modeling techniques.
The method is simple, easy to reproduce and should serve as a stronger baseline for research work on unsupervised learning from videos.
It consists of a quantization step, followed by a convolutional extension of rNN. 
We evaluated the performance of this model on a relatively large video dataset showing that the model is able to generate
short sequences exhibiting non-trivial motion. 

This model shows that it is possible to learn the local spatio-temporal geometry of videos purely from data, 
without relying on explicit modeling of transformations. 
The temporal recurrence and spatial convolutions are key to regularize the estimation by indirectly assuming stationarity and locality. 
However, much is left to be understood. First, we have shown generation results that are valid only for short temporal intervals, 
after which long range interactions are lost. Extending the prediction to longer spatio-temporal intervals 
is an open challenge against the curse of dimensionality, which implies moving from pixel-wise predictions to more high-level 
features. Next, it remains to be seen whether the resulting features are useful to supervised tasks, such as action recognition.

{\bf ACKNOWLEDGEMENTS}
\newline
The authors would like to acknowledge Piotr Dollar for providing us the optical flow estimator, Manohar Paluri for his help
with the data, and all the FAIR team for insightful comments.

%% file: appendix.tex
\section{Supplementary Material}

\subsection{van Hateren's Dataset} \label{zebra}
\begin{figure}[!h]
\begin{center}
\includegraphics[width=0.3\linewidth]{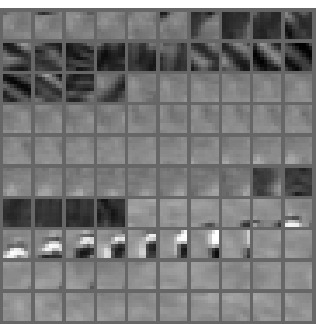}
\includegraphics[width=0.6\linewidth]{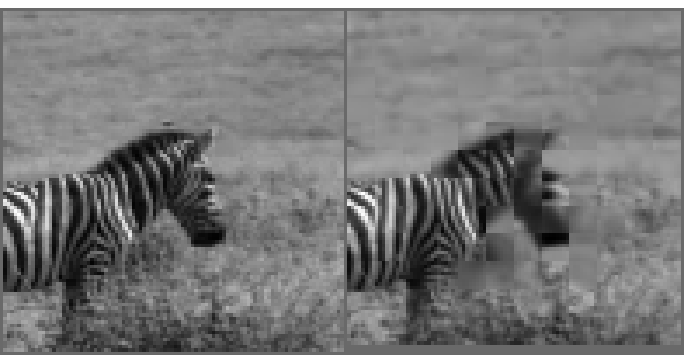}
\end{center}
\caption{van Hateren's video dataset: example of consecutive patches extracted at random spatial
locations (left) and example of a frame and its quantized version.}
\label{fig:vanHateren}
\end{figure}

The van Hateren dataset of natural videos has been a standard dataset for investigating
models of temporal sequences~\citep{cadieu-09, Olshausen-Field}.
Our version was downloaded from the material provided by~\citet{cadieu-09}
at~\url{https://github.com/cadieu/twolayer}. It consists of 56 videos, each 64
frames long. Each frame has size 128$\times$128 pixels. This is a very small dataset, where
objects are highly textured and move at similar speeds. Given the small dataset size, we could
only evaluate patch based models. Fig.~\ref{fig:vanHateren} shows examples of video patches
extracted from this dataset, along with the effect of quantization on the images.

Tab.~\ref{tab:logprob} (left)
compares n-grams (tri-grams worked worse than bi-grams and are therefore
omitted), neural net and rNN language models. Given the small dataset size, overfitting prevented
bigger models to perform better (on the validation/test sets).
We found that the neural net and the rNN work similarly and better than the bi-gram.

\begin{figure}[!h]
\begin{center}
\includegraphics[width=0.8\linewidth]{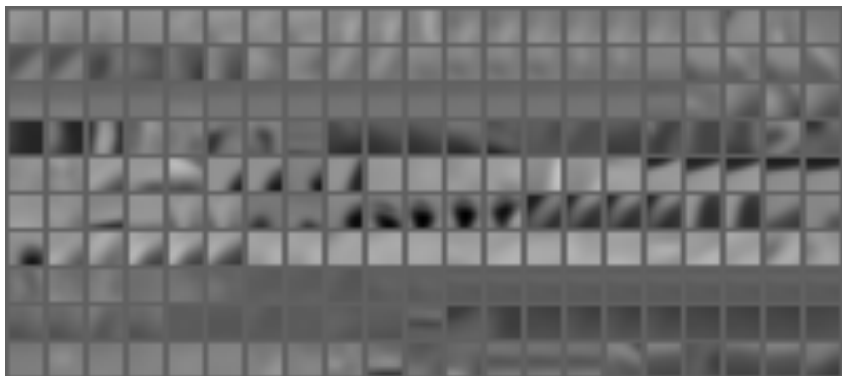}
\end{center}
\caption{Example of generation of 8$\times$8 pixel image patches after training on the van Hateren's video dataset.
Each row is an independent sequence. Columns are consecutive time steps (read from left to right).}
\label{fig:vanhaterenGeneration}
\end{figure}

\newpage
\subsection{Genration Examples}
Please, refer to \\
{\tiny\url{https://research.facebook.com/publications/video-language-modeling-a-baseline-for-generative-models-of-natural-videos/}}
for more examples.

\subsection{Filling Examples}
Here, we provide more examples and details about how we ran filling experiments.
The optical flow baseline of fig.~\ref{fig:filling_demo}, was computed by:
a) estimating the flow from two frames in the past, b) estimating the flow from two frames in the future, c) linearly interpolating 
the the flow for the missing frames and d) using the estimated flow to reconstruct the missing frames.

Below, you can find some examples of filled in frames; more examples are available at:\\
{\tiny\url{https://research.facebook.com/publications/video-language-modeling-a-baseline-for-generative-models-of-natural-videos/}}.

\begin{figure}[h]
\begin{center}
\vspace{-.2cm}
\includegraphics[width=0.8\linewidth]{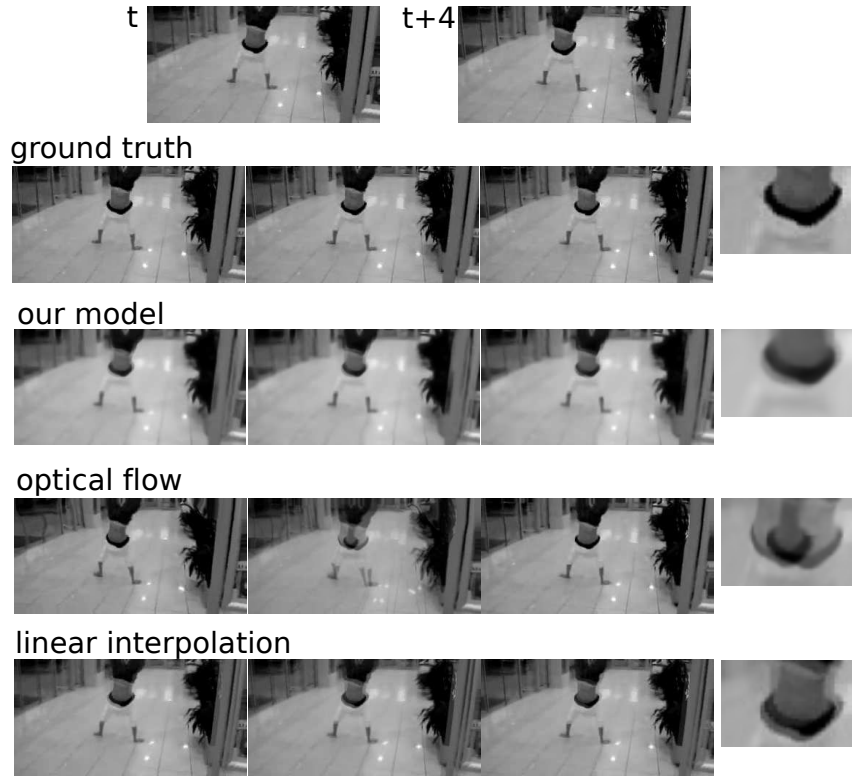}
\end{center}
\caption{Example of filling in missing frames. Top: the frames used for conditioning.
Second row: ground truth missing frames. Third row: our model. Fourth row: optical flow based
algorithm. Fifth row: linear interpolation. Last column: zoom of a patch from the missing middle frame.
See sec.~\ref{filling} for details.}
\label{fig:filling_demo1}
\end{figure}

\begin{figure}[h]
\begin{center}
\vspace{-.2cm}
\includegraphics[width=0.8\linewidth]{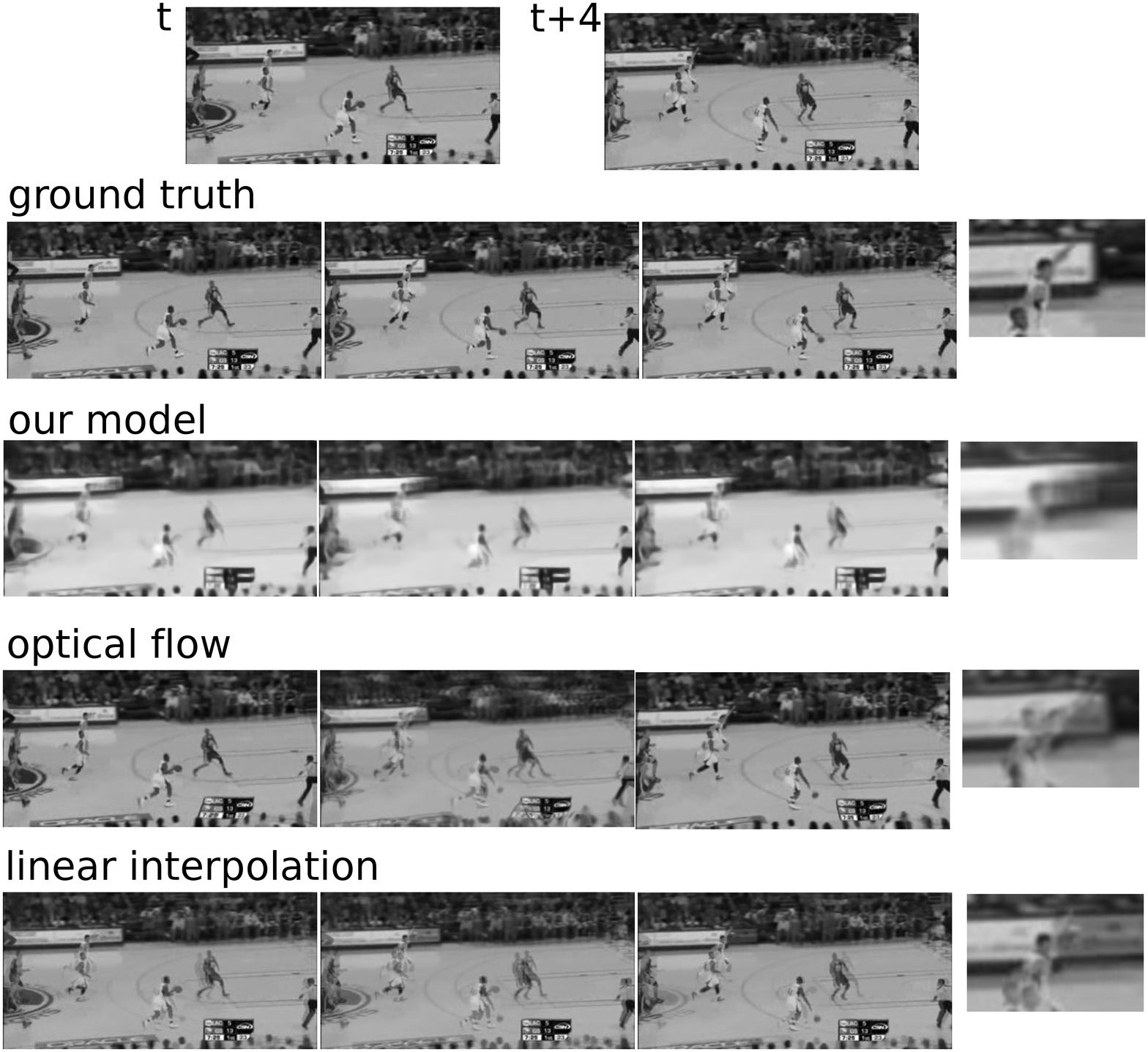}
\end{center}
\caption{Example of filling in missing frames. Top: the frames used for conditioning.
Second row: ground truth missing frames. Third row: our model. Fourth row: optical flow based
algorithm. Fifth row: linear interpolation. Last column: zoom of a patch from the missing middle frame.
See sec.~\ref{filling} for details.}
\label{fig:filling_demo2}
\end{figure}

\begin{figure}[h]
\begin{center}
\vspace{-.2cm}
\includegraphics[width=0.8\linewidth]{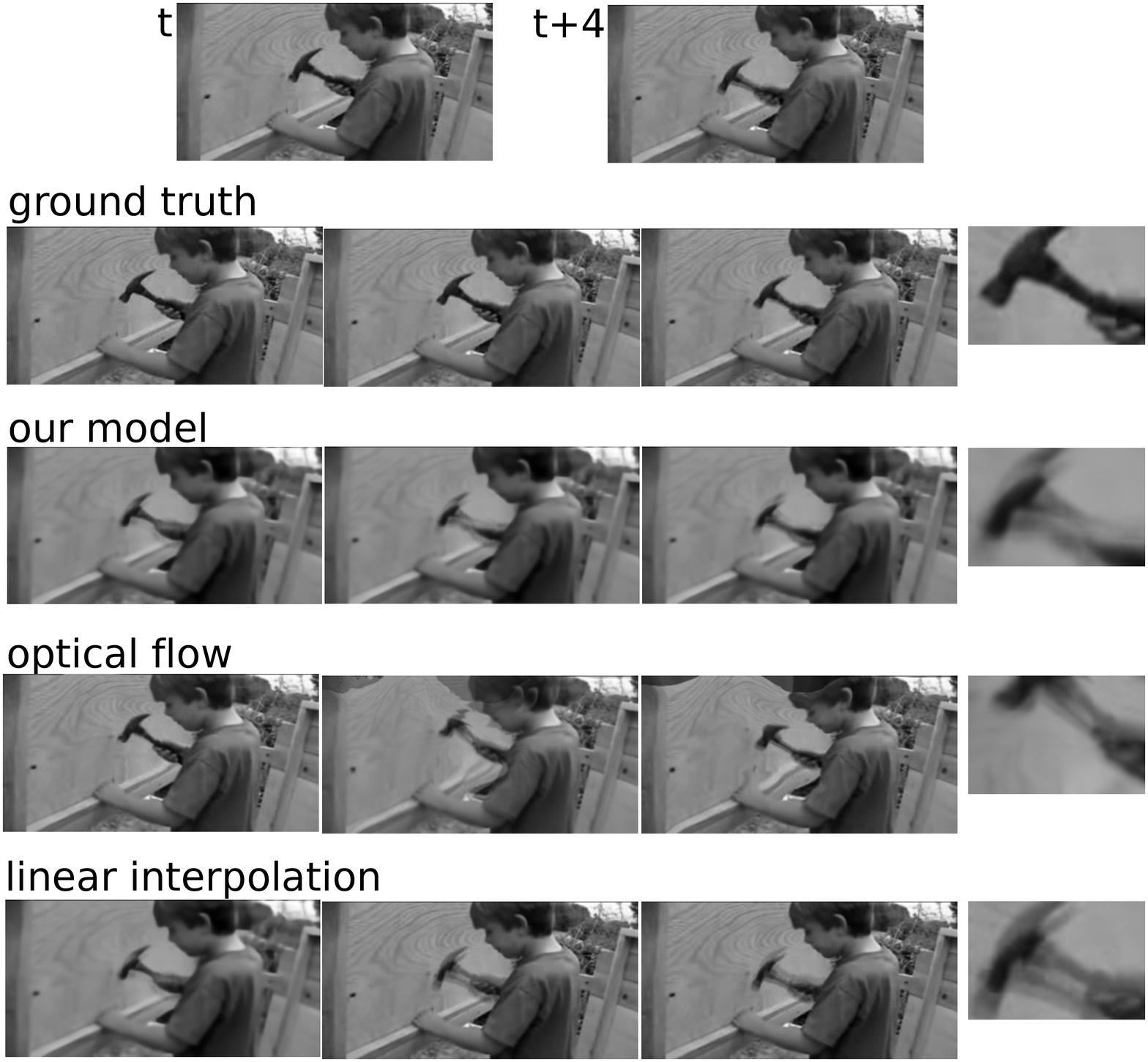}
\end{center}
\caption{Example of filling in missing frames. Top: the frames used for conditioning.
Second row: ground truth missing frames. Third row: our model. Fourth row: optical flow based
algorithm. Fifth row: linear interpolation. Last column: zoom of a patch from the missing middle frame.
See sec.~\ref{filling} for details.}
\label{fig:filling_demo3}
\end{figure}

\begin{figure}[h]
\begin{center}
\vspace{-.2cm}
\includegraphics[width=0.8\linewidth]{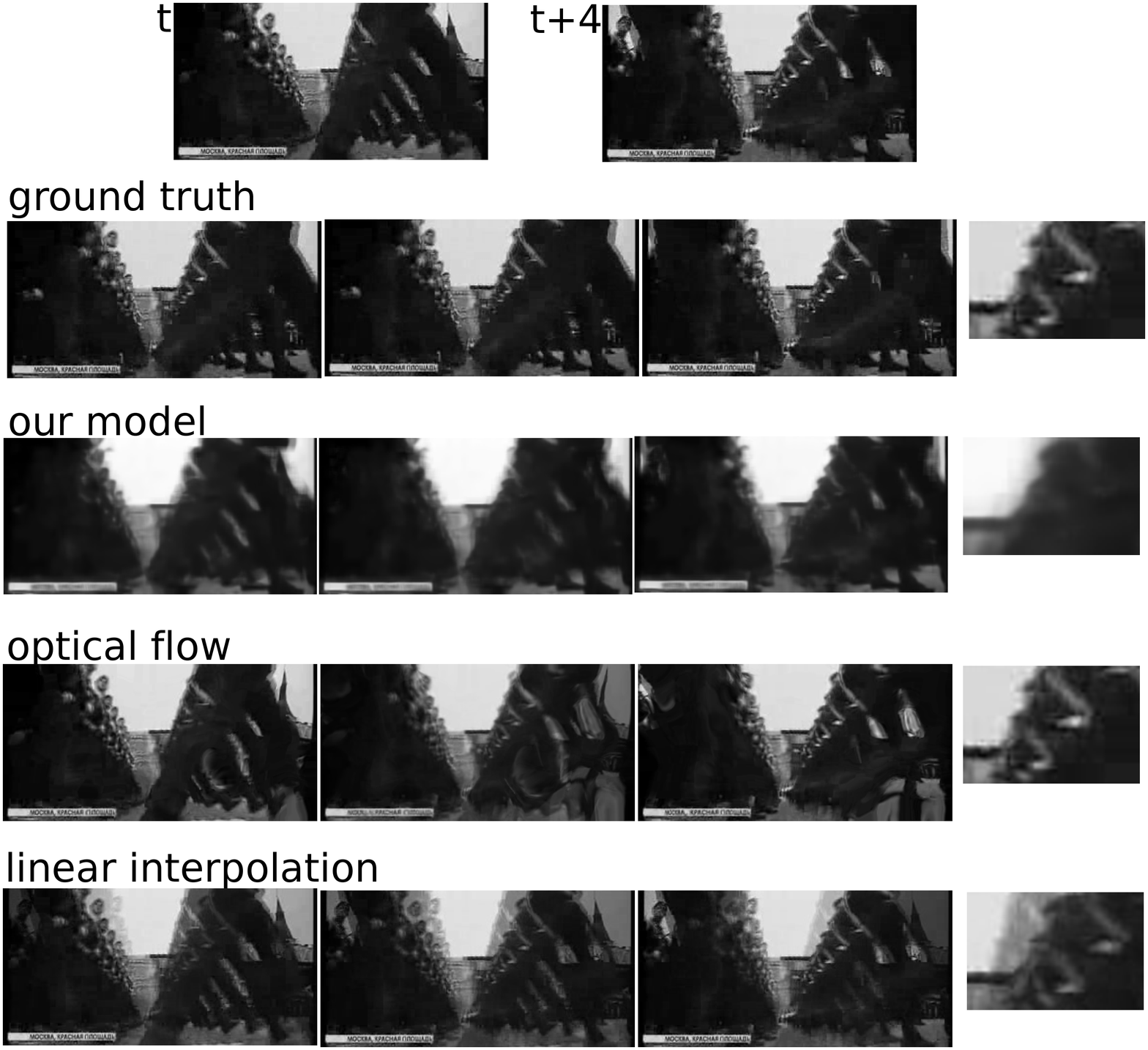}
\end{center}
\caption{Example of filling in missing frames. Top: the frames used for conditioning.
Second row: ground truth missing frames. Third row: our model. Fourth row: optical flow based
algorithm. Fifth row: linear interpolation. Last column: zoom of a patch from the missing middle frame.
See sec.~\ref{filling} for details.}
\label{fig:filling_demo4}
\end{figure}

%% file: videoLM_iclr2015.bbl
\begin{thebibliography}{32}
\providecommand{\natexlab}[1]{#1}
\providecommand{\url}[1]{\texttt{#1}}
\expandafter\ifx\csname urlstyle\endcsname\relax
  \providecommand{\doi}[1]{doi: #1}\else
  \providecommand{\doi}{doi: \begingroup \urlstyle{rm}\Url}\fi

\bibitem[Bengio et~al.(2003)Bengio, Ducharme, Vincent, and Jauvin]{nn-lm}
Bengio, Y., Ducharme, R., Vincent, P., and Jauvin, C.
\newblock A neural probabilistic language model.
\newblock \emph{JMLR}, 2003.

\bibitem[Brox et~al.(2004)Brox, Bruhn, Papenberg, and Weickert]{brox2004high}
Brox, T., Bruhn, A., Papenberg, N., and Weickert, J.
\newblock High accuracy optical flow estimation based on a theory for warping.
\newblock In \emph{Computer Vision-ECCV 2004}, pp.\  25--36. Springer, 2004.

\bibitem[Cadieu \& Olshausen(2009)Cadieu and Olshausen]{cadieu-09}
Cadieu, C. and Olshausen, B.
\newblock Learning transformational invariants from natural movies.
\newblock In \emph{NIPS}, 2009.

\bibitem[Girshick et~al.(2014)Girshick, Donahue, Darrell, and
  Malik]{girshick2014rcnn}
Girshick, Ross, Donahue, Jeff, Darrell, Trevor, and Malik, Jitendra.
\newblock Rich feature hierarchies for accurate object detection and semantic
  segmentation.
\newblock In \emph{Proceedings of the IEEE Conference on Computer Vision and
  Pattern Recognition ({CVPR})}, 2014.

\bibitem[Goodfellow et~al.(2014)Goodfellow, Pouget-Abadie, Mirza, Xu,
  Warde-Farley, Ozair, Courville, and Bengio]{goodfellow-nips2014}
Goodfellow, I, Pouget-Abadie, J., Mirza, M., Xu, B., Warde-Farley, D., Ozair,
  S., Courville, A., and Bengio, Y.
\newblock Generative adversarial nets.
\newblock In \emph{NIPS}, 2014.

\bibitem[Gregor \& LeCun(2010)Gregor and LeCun]{gregor_lc10}
Gregor, K. and LeCun, Y.
\newblock Emergence of complex-like cells in a temporal product network with
  local receptive fields.
\newblock arXiv:1006.0448, 2010.

\bibitem[Hariharan et~al.(2014)Hariharan, Arbel{\'a}ez, Girshick, and
  Malik]{hariharan-eccv2014}
Hariharan, B., Arbel{\'a}ez, P., Girshick, R., and Malik, J.
\newblock Simultaneous detection and segmentation.
\newblock In \emph{ECCV}, 2014.

\bibitem[Hinton \& Salakhutdinov(2006)Hinton and
  Salakhutdinov]{Hinton-DeepAutoencoder}
Hinton, G.E. and Salakhutdinov, R.~R.
\newblock Reducing the dimensionality of data with neural networks.
\newblock \emph{Science}, 2006.

\bibitem[Hurri \& Hyv{\"a}rinen(2003)Hurri and Hyv{\"a}rinen]{hurri03}
Hurri, J. and Hyv{\"a}rinen, A.
\newblock Simple-cell-like receptive fields maximize temporal coherence in
  natural video.
\newblock \emph{Neural Computation}, 2003.

\bibitem[Jojic \& Frey(2001)Jojic and Frey]{sprites01}
Jojic, N. and Frey, B.J.
\newblock Learning flexible sprites in video layers.
\newblock In \emph{CVPR}, 2001.

\bibitem[Kavukcuoglu et~al.(2008)Kavukcuoglu, Ranzato, and LeCun]{psd}
Kavukcuoglu, K., Ranzato, M., and LeCun, Y.
\newblock Fast inference in sparse coding algorithms with applications to
  object recognition.
\newblock ArXiv 1010.3467, 2008.

\bibitem[Krizhevsky et~al.(2012)Krizhevsky, Sutskever, and
  Hinton]{Krizhevsky:NIPS12}
Krizhevsky, A., Sutskever, I., and Hinton, G.
\newblock Image{N}et classification with deep convolutional neural networks.
\newblock In \emph{NIPS}, 2012.

\bibitem[Lee et~al.(2007)Lee, Chaitanya, and Ng]{lee-nips-07}
Lee, H., Chaitanya, E., and Ng, A.~Y.
\newblock Sparse deep belief net model for visual area v2.
\newblock In \emph{Advances in Neural Information Processing Systems}, 2007.

\bibitem[Memisevic \& Hinton(2009)Memisevic and Hinton]{MemisevicHinton09}
Memisevic, R. and Hinton, G.E.
\newblock Learning to represent spatial transformations with factored
  higher-order boltzmann machines.
\newblock \emph{Neural Computation}, 22:\penalty0 1473--1492, 2009.

\bibitem[Miao \& Rao(2007)Miao and Rao]{rao-2007}
Miao, X. and Rao, R.
\newblock Learning the lie groups of visual invariance.
\newblock \emph{Neural Computation}, 2007.

\bibitem[Michalski et~al.(2014)Michalski, Memisevic, and
  Konda]{memisevic-nips2014}
Michalski, V., Memisevic, R., and Konda, K.
\newblock Modeling deep temporal dependencies with recurrent ''grammar cells''.
\newblock In \emph{NIPS}, 2014.

\bibitem[Mikolov et~al.(2010)Mikolov, Karafiat, Burget, Cernocky, and
  Khudanpur]{rnn-lmm}
Mikolov, T., Karafiat, M., Burget, L., Cernocky, J., and Khudanpur, S.
\newblock Recurrent neural network based language model.
\newblock In \emph{Proc.Interspeech}, 2010.

\bibitem[Olshausen \& Field(1997)Olshausen and Field]{Olshausen-Field}
Olshausen, B.~A. and Field, D.~J.
\newblock Sparse coding with an overcomplete basis set: a strategy employed by
  v1?
\newblock \emph{Vision Research}, 37:\penalty0 3311--3325, 1997.

\bibitem[Ostrovsky et~al.(2009)Ostrovsky, Meyers, Ganesh, Mathur, and
  Sinha]{ostrovsky-2009}
Ostrovsky, Y., Meyers, E., Ganesh, S., Mathur, U., and Sinha, P.
\newblock Visual parsing after recovery from blindness.
\newblock \emph{Psychological Science}, 2009.

\bibitem[Pachitariu \& Sahani(2012)Pachitariu and Sahani]{sahani-nips12}
Pachitariu, M. and Sahani, M.
\newblock Learning visual motion in recurrent neural networks.
\newblock In \emph{NIPS}, 2012.

\bibitem[Ranzato et~al.(2013)Ranzato, Mnih, Susskind, and
  Hinton]{ranzato-pami2013}
Ranzato, M., Mnih, V., Susskind, J., and Hinton, G.
\newblock Modeling natural images using gated mrfs.
\newblock \emph{PAMI}, 2013.

\bibitem[Rumelhart et~al.(1986)Rumelhart, Hinton, and Williams]{autoencoder-nn}
Rumelhart, D.E., Hinton, G.E., and Williams, R.J.
\newblock Learning representations by back-propagating errors.
\newblock \emph{Nature}, 323:\penalty0 533--536, 1986.

\bibitem[Simonyan \& Zisserman(2014)Simonyan and Zisserman]{simonyan-nips2014}
Simonyan, K. and Zisserman, A.
\newblock Two-stream convolutional networks for action recognition in videos.
\newblock In \emph{NIPS}, 2014.

\bibitem[Soomro et~al.(2012)Soomro, Zamir, and Shah]{ucf101}
Soomro, K., Zamir, A.R., and Shah, M.
\newblock Ucf101: A dataset of 101 human action classes from videos in the
  wild.
\newblock CRCV-TR-12-01, 2012.

\bibitem[Sutskever et~al.(2009)Sutskever, Hinton, and Taylor]{sutskever09}
Sutskever, I., Hinton, G.E., and Taylor, G.W.
\newblock The recurrent temporal restricted boltzmann machine.
\newblock In \emph{NIPS}, 2009.

\bibitem[Taylor et~al.(2011)Taylor, Hinton, and Roweis]{taylor-jmlr11}
Taylor, G.W., Hinton, G.E., and Roweis, S.~T.
\newblock Two distributed-state models for generating high-dimensional time
  series.
\newblock \emph{JMLR}, 2011.

\bibitem[Theis et~al.(2011)Theis, Gerwinn, Sinz, and Bethge]{theis-2011}
Theis, L., Gerwinn, S., Sinz, F., and Bethge, M.
\newblock In all likelihood, deep belief is not enough.
\newblock \emph{JMLR}, 2011.

\bibitem[Theis et~al.(2012)Theis, Hosseini, and Bethge]{Theis2012a}
Theis, L., Hosseini, R., and Bethge, M.
\newblock Mixtures of conditional gaussian scale mixtures applied to multiscale
  image representations.
\newblock \emph{PLoS ONE}, 7\penalty0 (7), 2012.

\bibitem[van Hateren \& Ruderman(1998)van Hateren and Ruderman]{hateren98}
van Hateren, J.H. and Ruderman, D.L.
\newblock Independent component analysis of natural image sequences yields
  spatio-temporal filters similar to simple cells in primary visual cortex.
\newblock \emph{Royal Society}, 1998.

\bibitem[Vincent et~al.(2008)Vincent, Larochelle, Bengio, and
  Manzagol]{vincent-icml08}
Vincent, P., Larochelle, H., Bengio, Y., and Manzagol, P.A.
\newblock Extracting and composing robust features with denoising autoencoders.
\newblock In \emph{ICML}, 2008.

\bibitem[Wiskott \& Sejnowski(2002)Wiskott and Sejnowski]{SFA}
Wiskott, L. and Sejnowski, T.
\newblock Slow feature analysis: unsupervised learning of invariances.
\newblock \emph{Neural Computation}, 14\penalty0 (4):\penalty0 715--770, 2002.

\bibitem[Zoran \& Weiss(2012)Zoran and Weiss]{zoran-2012}
Zoran, D. and Weiss, Y.
\newblock Natural images, gaussian mixtures and dead leaves.
\newblock In \emph{NIPS}, 2012.

\end{thebibliography}
